\pdfoutput=1 

\documentclass{article}

     \usepackage[final,nonatbib]{neurips_2019}

\usepackage[utf8]{inputenc} 
\usepackage[T1]{fontenc}    
\usepackage{hyperref}       
\usepackage{url}            
\usepackage{booktabs}       
\usepackage{amsfonts}       
\usepackage{nicefrac}       
\usepackage{microtype}      
\usepackage{amsmath}
\usepackage{graphicx,color}
\usepackage{epsfig}
\usepackage{amsmath,amsthm,amssymb}
\usepackage{multirow}

\newtheorem{defn}{Definition}

\theoremstyle{remark}

\graphicspath{{figures/}}

\usepackage{subcaption}
\usepackage{algorithm}
\usepackage{algorithmic}

\title{On Higher-order Moments in Adam}

\author{%
  Zhanhong Jiang\\
  Johnson Controls International\\
  \texttt{zhanhong.jiang@jci.com}
  \And
  Aditya Balu\\
  Iowa State University\\
  \texttt{baditya@iastate.edu} \\
  \And
  Sin Yong Tan\\
  Iowa State University\\
  \texttt{tsyong98@iastate.edu}\\
  \And
  Young M Lee \\
  Johnson Controls International\\
  \texttt{young.m.lee@jci.com}\\
  \And
  Chinmay Hegde \\
  New York University \\
  \texttt{chinmay@iastate.edu}
  \\
  \And
  Soumik Sarkar\\
  Iowa State University\\
  \texttt{soumiks@iastate.edu} \\
}

\begin{document}

\maketitle

\begin{abstract}
In this paper, we investigate the popular deep learning optimization routine, Adam, from the perspective of statistical moments. While Adam is an adaptive lower-order moment based (of the stochastic gradient) method, we propose an extension namely, HAdam, which uses higher order moments of the stochastic gradient. Our analysis and experiments reveal that certain higher-order moments of the stochastic gradient are able to achieve better performance compared to the vanilla Adam algorithm. We also provide some analysis of HAdam related to odd and even moments to explain some intriguing and seemingly non-intuitive empirical results. 
\end{abstract}

\section{Introduction}\label{intro}
Adaptive gradient descent methods have become very popular in the optimization methods for machine learning, especially deep learning. Adagrad~\cite{duchi2011adaptive}, Adadelta~\cite{zeiler2012adadelta}, RMSprop~\cite{tieleman2012lecture}, and Adam~\cite{kingma2014adam} are the most popular optimizers that perform quite well in practice and hence have received considerable attention. While moments of stochastic gradients play a critical role in these optimizers, so far, only the first and second raw moments have been used in practice. To the best of our knowledge, the use of higher-order moments have not been reported for use in the aforementioned adaptive gradient descent methods; these could be beneficial for speeding up training of deep neural networks. 

This paper presents some interesting findings observed from using Adam with higher order moments. Although comprehensive theoretical analysis is not presented in this context, interesting findings arise. We hope these findings could provide some useful insights for the community to improve the understanding of these adaptive gradient descent methods and to develop more robust optimizers. Further, we show, some experimental evidences of better performance of the higher-order moments. This further reveals an interesting finding that odd-order moments lead to the divergence of Adam. It should also be noted that the higher-order moments can be applied to other adaptive gradient descent methods which have the similar form as Adam.
\section{Preliminaries}
\label{prelim}
Deep learning problems can be cast as optimization problems, which mostly are represented by an empirical risk minimization (ERM) as follows:
\begin{equation}
    \textnormal{min}\frac{1}{n}\sum^n_{i=1}f^i(x)
\end{equation}
where $n$ is the cardinality of a dataset, $f$ is the mapping from $\mathcal{R}^d\to\mathcal{R}$, and $x\in\mathcal{R}^d$ signifies the weight to be learned from data. Due to the broad applicability of ERM, 
for deep learning, $f$ is generically assumed to be non-convex and smooth, which is followed in this study.

Before we investigate the algorithm for minimizing $f$, we define formally several key concepts.

\begin{defn}\label{defn1}
The $k$-th moment of a real-valued continuous function $h(y)$ of a real variable $y$ about a value $c$ is defined as
\begin{equation}\label{k_moment}
    p_k=\int_{-\infty}^{\infty}(y-c)^kh(y)dy
\end{equation}
\end{defn}
Assume that $Y$ is a random variable. If $h(y)$ is a probability density function and $c=0$, then the $k$-th moment is the expected value of $Y^k$ and called a \textit{raw} moment.
Thus, Eq.~\ref{k_moment} can be converted into 
\begin{equation}\label{moment_zero}
    p'_k=\mathbb{E}(Y^k)=\int_{-\infty}^{\infty}y^kdH(y)
\end{equation}
where $H(y)$ is the cumulative probability distribution function. The first raw moment is the mean, denoted by $\mu=\mathbb{E}(Y)$. We now introduce an approximation of any $k$-th raw moment that could be useful in a discrete manner. 
\begin{defn}\label{defn3}
$\frac{1}{N}\sum^N_{j=1}Y^k_j$ is defined as the $k$-th raw sample moment, which can be used to estimate the $k$-th raw moment shown in Eq.~\ref{moment_zero} for all $k$, where $N$ is the number of samples drawn from a population.
\end{defn}

With the preliminary tools required, we now present the core update law for Adam:
\begin{align}
    m_t &= \beta_1m_{t-1} + (1-\beta_1)g_t\\
    V_t &= \beta_2V_{t-1} + (1-\beta_2)g^2_t\\
    x_t &= x_{t-1} - \alpha\frac{\sqrt{1-\beta_2^t}}{1-\beta_1^t}\frac{m_t}{\sqrt{V_t}+\epsilon}
    \label{eqn:adam}
\end{align}
where $\beta_1, \beta_2\in[0,1)$, $g_t$ is the stochastic gradient of $f$ at $x_t$, $\alpha$ is the step size, and $\epsilon$ is chosen as a small constant.

There has been a large body of work in the literature that investigates the convergence of Adam as well as how adaptive learning rate benefits the convergence. We refer the interested readers to~\cite{ruder2016overview,zhou2018convergence} for more details.

\section{High-order Moment in Adam}
\label{high}

Using the core update law of Adam as shown in Eq. \ref{eqn:adam}, it can be rewritten using induction as following (assuming that $m_0$ and $V_0$ are both initialized as 0 in this context for simplicity.):
\begin{align}
    m_t &= (1-\beta_1)\sum^{t-1}_{i=0}\beta_1^ig_{t-i}\\
    V_t &= (1-\beta_2)\sum^{t-1}_{i=0}\beta_2^ig_{t-i}^2\label{eqn:induction}
\end{align}
It can be observed that $m_t$ and $V_t$ are different from Definition~\ref{defn3} due to the usage of exponential moving average. In~\cite{kingma2014adam}, the authors called $m_t$ and $V_t$ biased raw sample moments and the third equation of Adam is able to correct the bias correspondingly. In this paper, we extend the second moment to high-order moment. The following algorithmic framework summarizes the High-order Adam (HAdam).
\begin{algorithm}[h]
    \caption{HAdam}
    \begin{algorithmic}[1]\label{hadam}
    \STATE {\textbf{Input}: $m_0=0, V_0=0, x_0, \alpha, \epsilon, \beta_1, \beta_2, k=2d, d=\{1,2,...\}$}
    \FOR{$t=1:M$}
      \STATE {Calculate the stochastic gradient $g_t$}
      \STATE {$m_t = \beta_1 m_{t-1} + (1-\beta_1) g_t$}
      \STATE {$V_t = \beta_2 V_{t-1} + (1-\beta_2) g^k_t$}
      \STATE{$x_t=x_{t-1} - \alpha\frac{\sqrt[k]{1-\beta_2^t}}{1-\beta_1^t}\frac{m_t}{\sqrt[k]{V_t}+\epsilon}$}
    \ENDFOR
\end{algorithmic}
\end{algorithm}

For analysis, we assume that $\epsilon=0$. In HAdam, Line 6 signifies the update law of $x_t$ which includes the bias correction, as done in \cite{kingma2014adam}. Using Equation~\ref{eqn:induction} and updating them for higher-order moments, we get
\begin{equation}\label{biased_update}
\mathbb{E}[V_t] =\mathbb{E}[ (1-\beta_2)\sum^{t-1}_{i=0}\beta_2^ig^k_{t-i}]=\mathbb{E}[g^k_t](1-\beta_2)\sum^{t-1}_{i=0}\beta_2^i+\zeta=\mathbb{E}[g^k_t](1-\beta^t_2)+\zeta,
\end{equation}
where $\zeta$ is the non-stationarity caused by the true $k$-th raw moment $\mathbb{E}[g^k_t]$. Essentially, $\zeta=0$ if $\mathbb{E}[g^k_t]$ is stationary or $\zeta$ can be arbitrarily small when $\beta_1$ is manually chosen such that stochastic gradients in the past are assigned with small weights. By dividing by the term $1-\beta_2^t$ on both sides in Eq.~\ref{biased_update}, the initialization biases associated with $V_t$ can be corrected. The effective step taken at time instant $t$ is $\Delta x_t=\alpha \frac{\hat{m}_t}{\sqrt[k]{\hat{V}_t}}$, where $\hat{m}_t = \frac{m_t}{1-\beta_1^t}$ and $\hat{V}_t = \frac{V_t}{1-\beta_2^t}$. It has been illustrated in \cite{kingma2014adam} that the effective magnitude of steps for each time instant taken can be approximately bounded by the stepsize, i.e., $|\Delta x_t|\lessapprox \alpha$ when the order is equal to 2, which corresponds to Adam. This bound mainly follows from $|\mathbb{E}[g]/\sqrt{\mathbb{E}[g^2]}|\leq 1$. 

We next investigate the third and fourth raw moments of the HAdam (i.e. $k = 3,4$) to obtain some insights. A detailed analysis of generic order moments will be deferred to an extended version of this work. 

The approach to study the third and fourth raw moments is to check the metric as follows
\begin{equation}
    \mathcal{M}_k = |\mathbb{E}[g]/\sqrt[k]{\mathbb{E}[g^k]}|
\end{equation}
In HAdam, we would also like to see if the bound $|\Delta x_t|\lessapprox \alpha$ holds true for higher orders.

\textbf{Fourth raw moment:} The metric defined earlier can be represented as $\mathcal{M}_4 = |\mathbb{E}[g]/\sqrt[4]{\mathbb{E}[g^4]}|$. As
\begin{equation}
    \mathbb{E}[g^4]=\mathbb{E}[(g^2)^2]=(\mathbb{E}[g^2])^2+Var(g^2),
\end{equation}
where $Var(\cdot)$ is the variance operator, then the following can be obtained 
\begin{equation}
    \mathbb{E}[g^4]=(Var(g))^2+2Var(g)(\mathbb{E}[g])^2+Var(g^2)+(\mathbb{E}[g])^4\geq (\mathbb{E}[g])^4
\end{equation}
The first equality follows from that $\mathbb{E}[g^2] = (\mathbb{E}[g])^2 + Var(g)$. Therefore, for the fourth raw moment, the metric $\mathcal{M}_k$ still holds true.

\textbf{Third raw moment:} A concept of \textit{skewness} is adopted in this context to help characterize the analysis. Denote by $\gamma$, the skewness of the stochastic gradients, which can be expressed as follows:
\begin{equation}
    \gamma = \frac{\mathbb{E}[g^3]-3\mathbb{E}[g]Var(g)-(\mathbb{E}[g])^3}{(\sqrt{Var(g)})^3},
\end{equation}
which yields that $\mathbb{E}[g^3]=\gamma(\sqrt{Var(g)})^3+3\mathbb{E}[g]Var(g)+(\mathbb{E}[g])^3$. The skewness value of stochastic gradients can be negative, positive, or undefined \cite{doane2011measuring}, and therefore the condition $\mathcal{M}_3\leq 1$ is not rigorously guaranteed. 

When $\mathcal{M}_3\gg 1$, the effective step taken could be significantly large, i.e., $|\Delta x_t|\gg \alpha$, resulting in the divergence. The upper bound for the effective step can also be understood as establishing a \textit{trust region} around the parameter value $x_t$ at time instant $t$ and the parameter space beyond this region cannot benefit from the current stochastic gradient. For HAdam with arbitrary even-orders, we can generalize the above analysis from the fourth raw moment such that $\mathcal{M}_k\leq 1, k=2d, d=\{1,2,...\}$. Thus, when $k\gg 2$, we have $|\mathbb{E}[g]/\sqrt[k]{\mathbb{E}[g^k]}\ll|\mathbb{E}[g]/\sqrt{\mathbb{E}[g^2]}|\leq 1$. 

\section{Numerical Results}\label{exp}

In this section, we show some preliminary results of HAdam on the CIFAR-10 dataset. We use a simple network architecture: we use two convolutional layers followed by a max pooling layer, a dropout layer, a dense layer, another dropout layer, and the final dense layer. The total number of learnable parameters is roughly 2,100,000. The default step size of $0.001$ is used and the number of epochs is 50. 

\begin{table}
  \caption{HAdam performance with different orders on Cifar 10 with 5 epochs}
  \label{hadam_cifar}
  \centering
  \resizebox{0.95\textwidth}{!}{
  \begin{tabular}{llllllllll}
    \toprule
    Order     & 2&3&4&5&6&7&8&9  \\
    \midrule
    Training Loss& 1.0323&nan&0.9508 &nan&0.9684&nan&0.9201&nan    \\
    Training Accuracy& 0.6343& 0.1000&0.6632 &0.1000&0.6588&0.1000&0.6761&0.1000      \\
    Test Loss&  1.0451&nan&0.9512&nan&0.9518&nan&0.9254&nan                       \\
    Test Accuracy& 0.6306 &0.1000&0.6705&0.1000&0.6703&0.1000&\textbf{0.6812}&0.1000           \\
    \bottomrule
  \end{tabular}
  }
\end{table}

Table~\ref{hadam_cifar} shows the performance of HAdam on the CIFAR-10 dataset with different orders of moment. The training and testing performance after 5 epochs is reported.  We observe that using odd-order moments in HAdam results in divergence of the algorithm, and that even-order moments have better performance than regular Adam. Since, higher-order raw sample moments should have played similar roles either for the odd or even moments, this interesting finding is somewhat counter-intuitive. However, the results could be explained from the analysis done for HAdam(order=3). At the same time, the even higher orders performs better. 

\begin{figure}[h!]
  \includegraphics[width=1.1\textwidth, trim={1.7in, 0.5in, 0.0in, 1.0in}, clip]{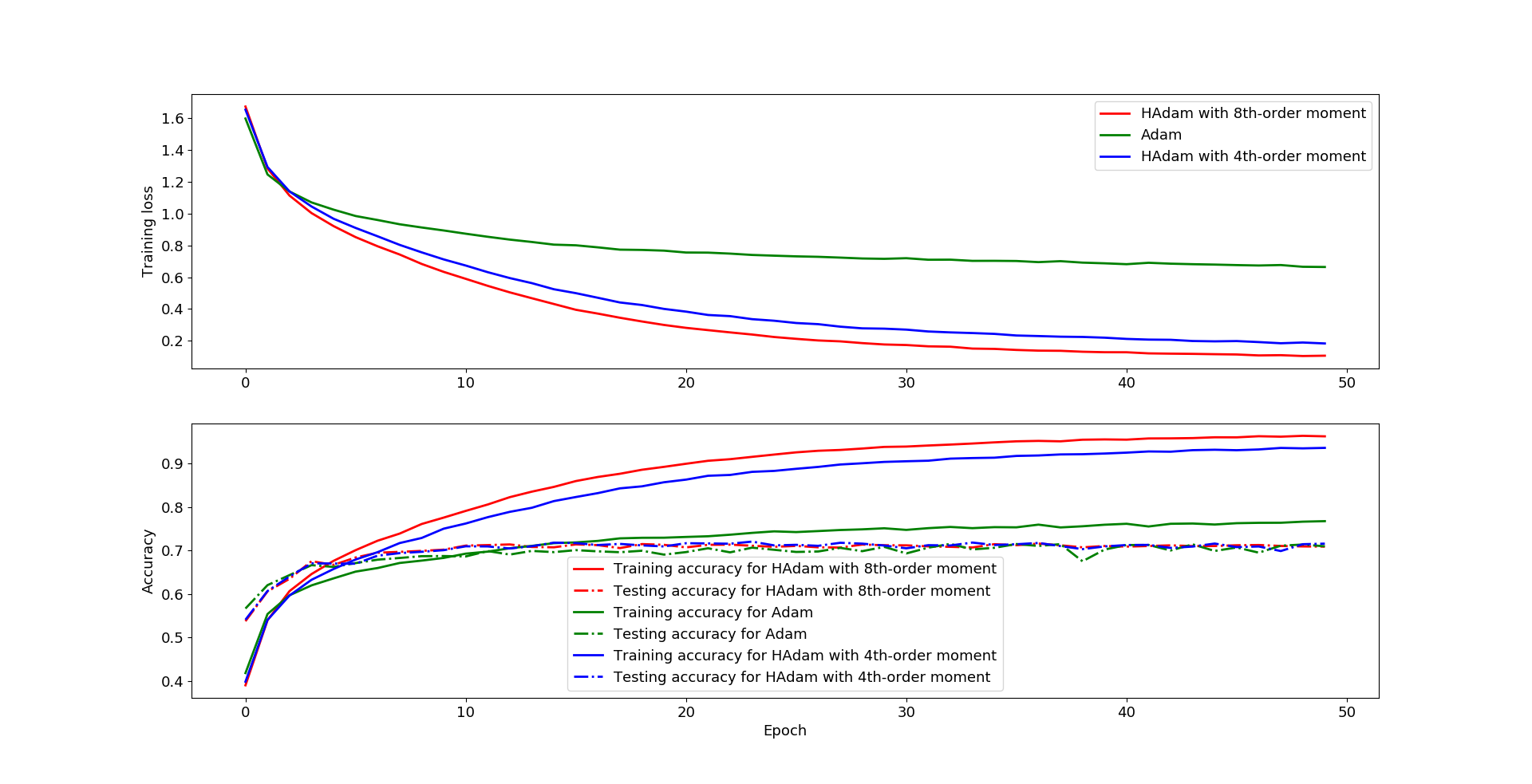}
    \centering
  \caption{Performance comparison between HAdam and Adam on Cifar 10}
  \label{cifar_exp}
\end{figure}

Figure~\ref{cifar_exp} depicts the performance comparison between HAdam and Adam. Two interesting observations can be made from Figure~\ref{cifar_exp}. The first one is in terms of training: HAdam performs better than Adam, which may be attributed to a better adaptive learning rate when the order is higher than 2. The second interesting finding is that in HAdam, higher order moments can lead to more overfitting, but still with quite similar testing performance as Adam. To summarize, HAdam allows for faster training and achieves quite similar testing performance compared with Adam. Beyond existing works, validation on models with more complex architectures will be addressed in future work.



\section{Conclusions}\label{conclusion}
This paper investigated the choice of moments in the Adam optimizer. Specifically, we found that odd-order moments could result in divergence of Adam, while even-order moment achieve better or similar performance as Adam. The convergence is faster and the training loss is better, but generalization could be worse than regular Adam. The results are also aligned with insights from our preliminary analysis which suggest why even-order moments could have convergence, and why odd-order moments could result in divergence.

\bibliographystyle{unsrt}
\bibliography{adam}
\clearpage

\end{document}